\documentclass[letterpaper, 10pt, conference]{ieeeconf}      
\usepackage{graphicx}
\usepackage{amsfonts}
\usepackage{amsmath}
\usepackage{csquotes}
\usepackage{subfig}
\usepackage{url}
\usepackage{booktabs}
\usepackage[vlined,algoruled,commentsnumbered]{algorithm2e} 
\usepackage[keeplastbox,spread,noautobase]{flushend}
\usepackage{multirow}
\usepackage{coverpage}

\IEEEoverridecommandlockouts                              

\newcommand{\etal}{et al.}
\let\oldnl\nl
\newcommand{\nonl}{\renewcommand{\nl}{\let\nl\oldnl}}

\newlength\lenKwIn
\newcommand\myKwIn[1]{%
  \settowidth\lenKwIn{\KwIn{}}%
  \setlength\hangindent{\lenKwIn}%
  \nonl\hspace*{\lenKwIn}#1\\}

\newlength\lenKwOut

\title{\LARGE \bf
On Multi-robot Search for a Stationary Object
}

\author{Miroslav Kulich$^{1}$, Libor P\v{r}eu\v{c}il$^{1}$ and  Juan Jos\'e Miranda Bront$^{2}$%
\thanks{The authors are with Czech Institute of Informatics, Robotics and Cybernetics, Czech Technical University in Prague,%
      Zikova 1903/4, 166 36 Prague, Czech Republic 
   {~~\tt\small kulich@cvut.cz,lhotsvoj@fel.cvut.cz, preucil@cvut.cz}%
}
}

\author{Miroslav Kulich$^{1}$, Libor P\v{r}eu\v{c}il$^{1}$ and  Juan Jos\'e Miranda Bront$^{2}$%
\thanks{$^{1}$ Miroslav Kulich, Libor P\v{r}eu\v{c}il are with Czech Institute of Informatics, Robotics and Cybernetics, Czech Technical University in Prague,%
      Zikova 1903/4, 166 36 Prague, Czech Republic%
{\tt\small ~\{miroslav.kulich, libor.preucil\}@cvut.cz}}%
\thanks{$^{2}$ Juan Jos\'e Miranda Bront is with Departamento de Computaci\'{o}n, Facultad de Ciencias Exactas y Naturales, Universidad de Buenos Aires, Argentina
{\tt\small ~jmiranda@dc.uba.ar}}%
}

\begin{document}
\mauthor{Miroslav Kulich, Libor P\v{r}eu\v{c}il and  Juan Jos\'e Miranda Bront}
\published{{\it Proceedings of the European Conference on Mobile Robotics (ECMR) 2017}, Paris. ISBN 978-1-5386-1096-1.}
\DOI{10.1109/ECMR.2017.8098696}
\original{https://ieeexplore.ieee.org/document/8098696}
\coverpage
\twocolumn

\maketitle
\thispagestyle{empty}
\pagestyle{empty}

\begin{abstract}
Two variants of multi-robot search for a stationary object in a priori known environment represented by a graph are studied in the paper. 
The first one is generalization of the Traveling Deliveryman Problem where more than one deliveryman is allowed to be used in a solution.   
Similarly, the second variant is generalization of the Graph Search Problem.
A novel heuristics suitable for both problems is proposed which is furthermore integrated into a cluster-first route second approach.  
A set of computational experiments was conducted over the benchmark instances derived from the TSPLIB library. 
The results obtained show that even a standalone heuristics significantly outperforms the standard solution based on k-means clustering in quality of results as well as computational time.  
The integrated approach furthermore improves solutions found by a standalone heuristics by up to 15\% at the expense of higher computational complexity.
\end{abstract}

\section{Introduction}
Assume a mobile robot autonomously operating in a priori known environment, in which a stationary object of interest is randomly placed. 
The objective is to find the object, whose position is not known in advance, in a minimal time.
This problem is formulated as the Traveling Deliveryman Problem (TDP) provided that the environment is represented by a graph and probability of appearance of the object is the same for all vertices in the graph.
Although TDP is similar to the Traveling Salesman Problem (TSP), objectives of these problems differ and thus an optimal solution for one problem is not necessarily optimal for the second problem~\cite{Kulich2014}.  

TDP, which is NP-hard~\cite{Afrati86}, has been studied from various perspectives during the last few years.
Besides exact algorithms introduced by several authors~\cite{Gouveia1995,Mendez-Diaz2008}, Integer Linear Programming with Branch-and-Cut and Branch-Cut-and-Price approaches were proposed~\cite{Abeledo2012,Miranda-Bront2013,Godinho2014} for time-dependent TSP which generalizes TDP. 
The best exact algorithm~\cite{Abeledo2012}, nevertheless, is able to solve instances with up to 107 vertices to optimality in several hours.

More useful are heuristics and meta-heuristics which provide good quality solutions with much lower computational effort. These rely particularly on Greedy Randomized Adaptive Search Procedure (GRASP), introduced originally by Feo and Resende~\cite{Feo1995}, and Variable Neighborhood Search (VNS), proposed by Hansen and Mladenovic~\cite{Hansen1997}. 
Salehipour \etal\ \cite{Salehipour2011} employ a GRASP for TDP and compare the impact of VNS procedure as a local search phase with Variable Neighborhood Descent. 
Mladenovic \etal\ \cite{Mladenovic2012} propose a General VNS (GVNS), which improves Salehipour's results. Further improvements were achieved by Silva \etal~\cite{Silva2012} who propose a simple multi-start heuristic combined with an Iterated Local Search procedure.
To the best of our knowledge, the approach by Silva is thus the one producing the best results, nevertheless, time needed to compute problems containing 100 vertices is more than ten seconds and instances with 200 vertices are computed in approximately one minute. 

Approaches used by the robotics community are simpler. 
Sarmiento \etal~\cite{SarmientoMH04} propose a modification of breadth-first algorithm which iteratively constructs all possible routes of the defined length, fixes the most promising one and starts the next search from this route as a prefix. 
Finally, a modified depth-first search algorithm with pruning and limited branching was introduced in our previous work~\cite{Kulich2014}.

The Graph Search Problem (GSP), introduced in Kutsoupias \etal\ \cite{Kutsoupias96}, is formulated under the same settings as TDP, with the only difference that each vertex has assigned probability of finding the object when visiting the vertex and probabilities of the vertices differ in general.
Besides some theoretical results regarding approximation schemes presented in Ausiello \etal\ \cite{Ausiello00}, no further developments are present in the related literature.
The only exception is our recent work~\cite{Kulich2016}, in which a tailored GRASP meta-heuristic for the GSP is introduced which is able to find near-optimal solutions for TDP and GSP problems up to 107 vertices in about one second of computing time.

Little attention has been paid to multi-robot variants of TDP and GSP. 
On the other hand, some inspiration can be found in approaches to the Multiple Traveling Salesman Problem (mTSP) or other routing problems.
Besides genetic algorithms, neural networks, or ant colony optimization, the cluster-first route-second approach plays an important role. 
The key idea of this approach is to split all vertices into $M$ clusters based on their location in space ($M$ is the number of salesmen) and solve the traditional Traveling Salesman Problem for each cluster separately.
For example, Sathyan \etal~\cite{Sathyan2015} use k-means clustering for the first phase followed by application of a genetic algorithm or 2-opt heuristic. 
Boone \etal~\cite{Boone2015} employ an initial k-means clustering and modify it by taking points from the cluster with the largest tour distance and adding them to one of the smaller clusters. Convex hulls, fuzzy logic, and the TSP solution are used to determine which points to switch.
Geetha \etal~\cite{geetha2009improved} improved k-means algorithm for the Capacitated Clustering Problem by incorporating a priority measure to the criterion on which are vertices assigned to clusters. 
We use k-means clustering followed by solution of Traveling Salesman Problem for each cluster by Chained Lin-Kernighan as a base of a goal assignment strategy in multi-robot exploration~\cite{Faigl2012}.
The presented experimental results indicate that this method provides more efficient assignment than former approaches.

In this paper we build on experience from the works mentioned above and present a cluster-first route-second approach for the Multiple Traveling Deliveryman Problem (mTDP) and the multi-vehicle case of GSP (mGSP). 
The approach extends our GRASP-based meta-heuristic for the single-vehicle problems~\cite{Kulich2016} by incorporating the clustering phase. 
The key contribution thus lies in design of a novel clustering method which respects special aspects of mTDP and mGSP.
The proposed solution is evaluated computationally and compared with an  algorithm based on k-means. 
The experimental results show that our solution has potential to be applied in practice as it provides better results than the k-means based approach in almost all problem instances.  

The rest of the paper is organized as follows. 
The problem is defined in Section~\ref{sec:problem}, the proposed clustering is presented in Section~\ref{sec:clustering}, while the key ideas of the tailored GRASP-approach for TDP and GSP are summarized in Section~\ref{sec:grasp}.
Computational results including instances of both mTDP and mGSP are presented and discussed in~\ref{sec:results}.
Finally, Section~\ref{sec:conclusion} is dedicated to concluding remarks and future directions.

\section{Problem Formulation}
\label{sec:problem}
That is, formally, given
\begin{itemize}
\item a complete undirected graph $G = (V,E)$, where $V= \{v_1,v_2,\dots,v_N\}$ stands for a finite set of vertices and $E$ is a set of edges between these vertices: $E=\{e_1,e_2, \dots e_{n^2}\}$, $e_{ij}=(v_i,v_j)$, $v_i,v_j\in V$, $i \ne j$,
\item $t:~E \rightarrow \mathbb{R}$: a cost $t_{ij}$ associated with each edge $e_{ij}$ representing time needed to traverse the shortest path from $i$ to $j$,
\item $p:~V \rightarrow \langle 0,1\rangle$: a weight for each vertex representing probability of presence of the searched object at the vertex,
\item the number of the vehicles $M$, and
\item $s_i \in V, i \in \left<1,M\right>$: starting vertices of the vehicles (note that several vehicles can start from the same vertex in general).   
\end{itemize}
\noindent Define a {\em walk} ${\boldsymbol \omega}=\langle \omega_1, \omega_2, \dots \omega_k\rangle$ as a sequence of vertices of $G$, i.e. $\omega_i\in V $ for $i\in\left<1,k\right>$.
The overall objective is then to find a tuple of $M$ walks $\Omega = \langle\boldsymbol\omega^1,\boldsymbol\omega^2,\dots, \boldsymbol\omega^M\rangle$
that visits all vertices of $V$ at least once (i.e. $\forall v\in V~\exists \omega^i_j\in {\boldsymbol \omega^i}:~ \omega^i_j=v$) and which minimizes the expected time to find the object:
\begin{equation}
{\Omega} = \arg\min_{W \in {\Theta}}\mathbb{E}(T|{W}) = \sum_{i=1}^M\sum_{j=1}^{|W^i|}\!\tau(w^i,j) p(w^i_j), 
\label{crit:search}
\end{equation}
where  
$$
\tau(w^i,j) = \sum_{\iota=1}^{j-1}t(w_\iota w^i_{\iota+1})
$$
is the time when the vertex $w^i_j$, the j-th vertex of the walk $W^i\in W$ is visited and $\Theta$ is a set of all possible sets of walks in $G$.
Moreover, all the walks have to start in the given starting vertices, i.e. $w^i_1 = s_i$ for $i=1\dots M$.
The minimal expected time is then
\begin{equation}
T_{exp}=\mathbb{E}(T|{\boldsymbol \omega}) = \sum_{i=1}^M\sum_{j=1}^{|\omega^i|}\!\tau(\omega^i,j) p(\omega^i_j).
\label{eq:texp}
\end{equation}

In summary, the aim is to minimize the average time the vertices are visited weighted by probabilities assigned to the vertices.

The multi-vehicle Traveling Deliveryman Problem is a special variant of mGSP with the only difference that the probability of finding the
object is the same for all the vertices in the graph. 
These probabilities can be omitted from the equations for this case.

\section{Proposed Clustering Approach}
\label{sec:clustering}

The proposed method for solving both mTDP and mGSP follows the cluster-first route-second schema.
This means that all vertices of the graph are divided into $M$ clusters assigned to the vehicles in the first phase.
A TDP/GSP solver is then run for each cluster separately to optimize the order in which the vertices in the clusters are visited.

The clustering approach whose scheme is depicted in Algorithm~\ref{alg:clustering} has a greedy nature.
The algorithm starts with initialization of clusters: each cluster contains the starting vertex of the vehicle it is assigned to and current time needed to traverse each route is set to zero (lines~\ref{l1}--\ref{l3}).
Moreover, the set of all not assigned graph vertices $V_{REST}$ is set (line~\ref{l4}). 

The main part of the algorithm is a loop between lines~\ref{l5}  and \ref{l18} which is processed until $V_{REST}$ is empty.
In each iteration of the loop, each not yet assigned vertex is examined to attach to each of the routes.
Time in which the vertex is visited is computed first (line~\ref{l9}, followed by evaluation of a penalty function (line~\ref{l10}).
This function was designed to prefer vertices that are visited earlier and with higher probability of the object of interest at them.

\begin{figure}[htb]
\centering
\includegraphics[width=0.8\columnwidth]{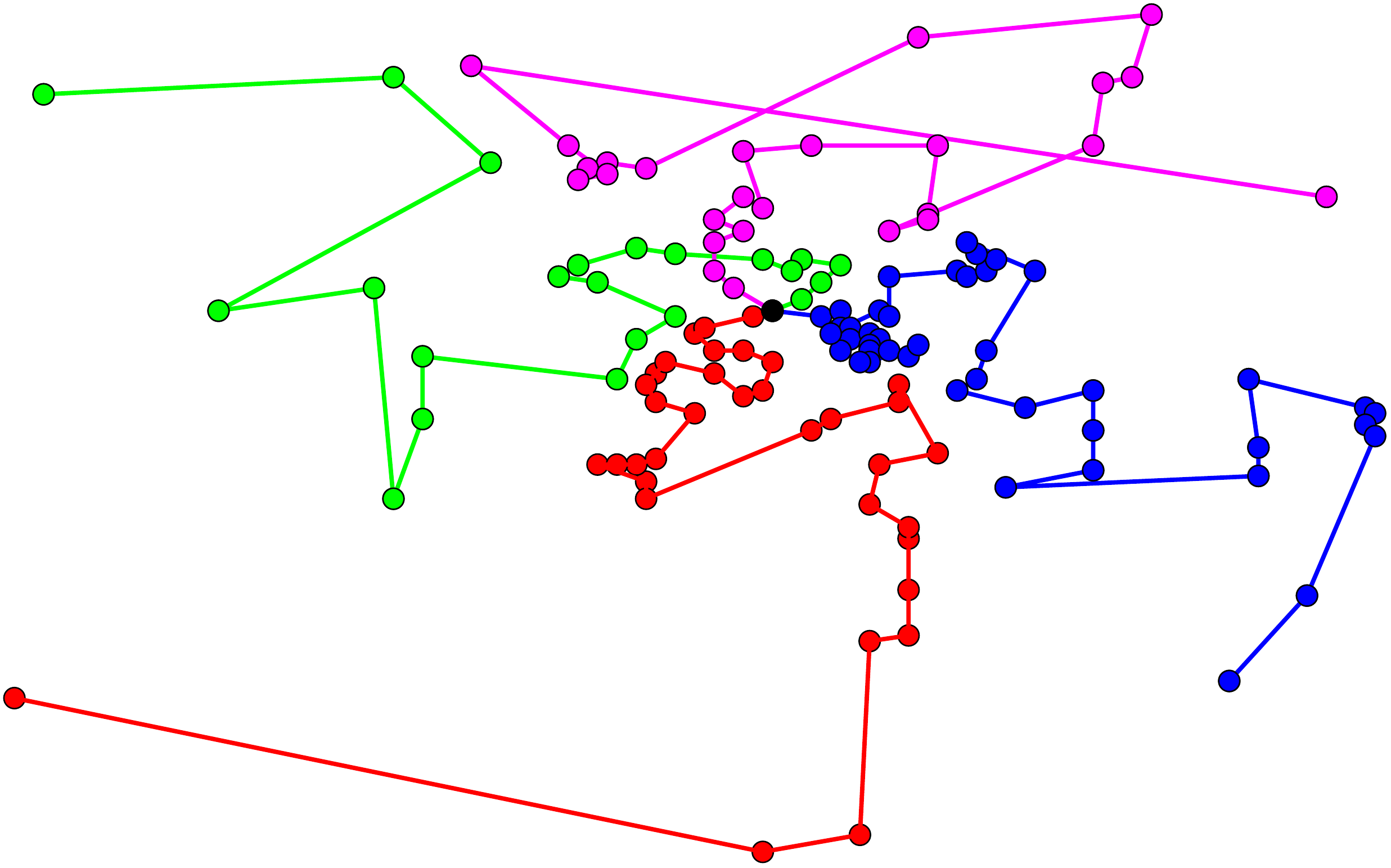}
\caption{The result of the proposed clustering method for the { bier127} problem~\cite{Reinelt91} and 4 vehicles.}
\label{fig:bier127}
\end{figure}

The pair $\langle v^{min}, \omega^{min}\rangle$ with the lowest value of the penalty function is chosen and the vertex $v^{min}$ is attached to the end of $\omega^{min}$ (line~\ref{l13}) and time needed to traverse the $\omega^{min}$ is updated accordingly (line~\ref{l14}).
Finally, $v$ is removed from $V_{REST}$ (line~\ref{l15}).
Note that the algorithm does not only return partition of vertices into clusters, but also their orders within the clusters (line~\ref{l19}). 
The result can be thus used as a solution of mTDP/mGSP.
An example of a solution found by the algorithm is shown in Fig.~\ref{fig:bier127}.

\LinesNumbered
\SetKwFor{Times}{}{times do}{end}
\DontPrintSemicolon
\begin{algorithm}[ht]
\KwIn{$M$ -- the number of vehicles}
\myKwIn{$G=(V,E)$\hspace{1em}-- a graph}
\myKwIn{$t_{ij}$\hspace{1em}-- costs of edges}
\myKwIn{$p(i)$\hspace{1em}-- probabilities associated with vertices}
\myKwIn{$s_i \in V, i \in \left<1,M\right>$\hspace{1em}-- start vertices of vehicles}
\KwOut{${\boldsymbol \omega}=\langle \omega_1, \omega_2, \dots \omega_M\rangle$ -- a tuple of sequences representing clusters}
\nonl\hrulefill\\

\For{$i\leftarrow 1$ \KwTo $M$}{\nllabel{l1}
  $\omega^i\leftarrow \langle s_i\rangle$\\\nllabel{l2} 
  $\tau_{\omega^i} = 0$\nllabel{l3}
}
$V_{REST} \leftarrow V\setminus\{s_1, s_2, \dots, s_M\}$\\\nllabel{l4}
\While{$V_{REST} \neq \emptyset$}{\nllabel{l5}
$min \leftarrow \infty$\\ \nllabel{l6}
\ForEach{$\omega \in \boldsymbol \omega$} {\nllabel{l7}
  \ForEach{$v \in V_{REST}$} {\nllabel{l8}
    $d = \tau_{\omega} + t(idx(v), idx(last(\omega)))$, where \\\nllabel{l9}
    \nonl\hspace{1em}$idx(v)$ is the index of $v$ and\\
    \nonl\hspace{1em}$last(\omega)$ is the last vertex of the route $\omega$\\
    $c = \frac{d}{1+p(idx(v))}$\\\nllabel{l10}
    \If{$c < min$}{\nllabel{l11}
        $min \leftarrow c$\\\nllabel{l12}
        $d^{min} \leftarrow d$\\\nllabel{l13}
        $v^{min} \leftarrow v$\\\nllabel{l14}
        $\omega^{min} \leftarrow \omega$\\ \nllabel{l15} 
    }
  }

  $\omega^{min} \leftarrow \omega^{min} + v$ \\ \nllabel{l16}
  $\tau_{\omega^{min}} \leftarrow d^{min}$\\\nllabel{l17}
  $V_{REST} \leftarrow V_{REST} \setminus \{v\}$\\\nllabel{l18}
}
}
\Return ${\boldsymbol \omega}=\langle \omega_1, \omega_2, \dots \omega_M\rangle$\\\nllabel{l19}
\caption{Proposed clustering algorithm.}
\label{alg:clustering}
\end{algorithm}

\section{Meta-heuristic for a single-vehicle case}
\label{sec:grasp}
Having the vertices partitioned into clusters, the second step is to find the best order for each cluster.
To do that, we employ the greedy randomized adaptive search procedure (GRASP) meta-heuristic tailored for TDP and GSP~\cite{Kulich2016}.
The general scheme of the GRASP is shown in Algorithm \ref{alg:grasp}.

The value of the best already found solution is initially set to a high number in the algorithm (line~\ref{b1}).
The meta-heuristic then consecutively constructs initial solutions using some simple, typically greedy, heuristic (line~\ref{b4}).  
The obtained solution is improved by a sequence of local search steps (line~\ref{b51}) and its cost is updated accordingly (line~\ref{b52}). 
If the improved solution is better than the current best solution, the best solution and its cost are updated (lines~\ref{b6}--\ref{b8}).
After all initial solutions are processed, the best solution found is returned (line~\ref{b9}).
More detailed description of the particular steps of the GRASP follows in the next paragraphs.

\LinesNumbered
\begin{algorithm}
  \caption{GRASP scheme.\label{alg:grasp}}
  $z_{\textrm{best}} \leftarrow \infty$. \\\label{b1}  
  \ForEach{$h \in H$}{\label{b2}
    \For{$k = 1,\dots,N_{it}$} {\label{b3}
      Obtain a feasible route $\omega$ using $h$.\\\label{b4}
      Improve route $\omega$ by applying a local search step  \\\label{b51}
      Update cost $z$ of $\omega$.\\\label{b52}
      \If{$z < z_{\textrm{best}}$}{\label{b6}
        $\omega_{\textrm{best}} \leftarrow P$\\\label{b7}
        $z_{\textrm{best}} \leftarrow z$\label{b8}
      } 
    }
}
\Return{$\omega_{\textrm{best}}$}\\\label{b9}
\end{algorithm}%

We consider two heuristics for initial construction of routes, both using the same general greedy scheme as shown in Algorithm~\ref{alg:constructive}.
The algorithm starts with a route containing only the start vertex (line~\ref{c1}), consecutively finds the best vertex from vertices not yet connected according to the penalty function $f$.
The found vertex is then appended at the end of the route (line~\ref{c4}) and removed from the set of not connected vertices (line~\ref{c5}).

The difference between the two heuristics lies in the definition of $f$, which affects the selection of the next vertex to be visited (line 4). We consider the standard distance-based function, $f_{\textrm{dist}}(u,v) = t(u,v)$. In order to incorporate the weights for each candidate, we further consider the function $f_{\textrm{ratio}}(u,v) = t(uv)/(1+p(v))$. 
As both heuristics are deterministic, we employ Monte Carlo randomization to generate various initial solutions -- the lower the penalty function value for the vertex, the higher probability that the vertex is selected.

After obtaining an initial feasible solution, an improvement phase is performed. 
In our case, a Variable Neighborhood Descent (VND)~\cite{Salehipour2011} is used.
The idea of VND is simple: a neighborhood of the current solution is completely searched and the best neigbor replaces the solution.
This process is repeated until no improvement is possible.
A neighborhood of the route $\omega$ defined by an operator is a set of all routes which are formed from $\omega$ by application of this operator.   
Two  different neighborhoods are considered which are formed by two  local search operators:
\begin{itemize}
\item \textbf{Swap}: Select two vertices in the tour and swap them.
\item \textbf{2-opt}: Select two non-adjacent arcs and replace them by two new arcs, obtaining a new route as a result.
\end{itemize}

When the result obtained by VND with Swap and 2-opt is promising, i.e. its cost is less than 10\% of the current best solution, the LK-op operator is applied to this result.
LK-op is an adaptation of Lin-Kernighan operator~\cite{Lin1973} designed for TSP. It starts with a feasible solution and considers each edge sequentially as a seed for an improvement procedure. 
The procedure attempts to obtain an improved route by application of a sequence of 2-opt moves, not necessarily improving the current solution, with a limited backtracking. 
If such a better path is found, it is accepted as the new initial solution and the procedure is restarted again from the first edge. 
Otherwise, the algorithm moves to the next edge, until all edges have been considered as a seed.

As LK-op is computationally demanding, the search space is reduced in two ways.
First, a length of a sequence of performed 2-opt moves is limited to a maximum depth $\alpha$.
Moreover, not all edges are considered for each vertex, only limited number of shortest edges $\beta$ is evaluated instead.
Detailed description of the method as well as discussion about complexity of the particular steps can be found in~\cite{Kulich2016}.

\begin{algorithm}
\caption{General greedy scheme.\label{alg:constructive}}
\KwIn{$G=(V,E)$\hspace{1em}-- a graph}
\myKwIn{$t_{ij}$\hspace{1em}-- costs of edges}
\myKwIn{$p(i)$\hspace{1em}-- probabilities associated with vertices}
\myKwIn{$s$\hspace{1em}-- starting vertex}
\KwOut{$\omega=\langle \omega_1, \omega_2, \dots \omega_K\rangle$ -- a route}
\nonl\hrulefill\\
  $P \leftarrow s$ \\\label{c1}
  $V_{REST} \leftarrow V \setminus \{s\}$ \\\label{c2}
  \While{$V_{REST} \neq \emptyset$}{\label{c3}
     $v \leftarrow \arg \min_{u \in V_{REST}} f(last(\omega),u)$\\\label{c4} 
     $\omega \leftarrow  \omega + v$\label{c5}
  }

  Return path $P$.
\end{algorithm}%

\section{Results}
\label{sec:results}

Performance of the proposed approach has been evaluated for both mTDP and mGSP.
The experiments for mTDP were run on a set of standard instances from TSPLIB~\cite{Reinelt91} with sizes between 52 and 1002.
As there are no benchmark instances for mGSP, instances from TSPLIB were also used for which probabilities of vertices were generated randomly. 
To ensure repeatability of experiments, a generator from \url{random.org} was utilized for generating 10000 normally distributed random numbers between 1 and 10 and the string "2016-09-11" was set as a seed\footnote{We are ready to publish the generated sequence if the paper is accepted.}. 
Random numbers were assigned to vertices respecting the order, i.e. $i$-th vertex of an TSPLIB instance has assigned $i$-th random number. 
Moreover, the numbers were normalized so that the sum of probabilities of all vertices for an instance is 1.

Regarding the parameters of the method, we set $\alpha = 20$, limiting the size of sequences of 2-opt moves, $\beta=5$ for first 4 depths of backtracking, and $\beta=1$ for the rest. $N_{it}$ was set to 50, generating 100 initial solutions in total.
Start positions of all vehicles were set to the first vertex of the instance. 

The proposed cluster-first route-second approach was compared with pure clustering as proposed in Section~\ref{sec:clustering} and with  an approach combining k-means clustering and our GRASP meta-heuristic. 
More specifically, we employed k-means++ -- an augmentation of k-means which significantly improves both the speed and the accuracy of k-means~\cite{Arthur2007}.
We refer to  the methods as {\tt proposed}, {\tt clustering}, and {\tt k-means}.

All experiments were performed within the same computational environment: a~workstation with the Intel\textregistered Core i7-3770 CPU at 3.4~Ghz running Linux with the kernel 4.4.0.
The algorithms have been implemented in C++ and compiled by clang 3.8.1. with \enquote{-O2} flag.
50 runs were run for each setup consisting of an instance, a number of vehicles, and a method to provide statistically significant results.

The results for mTDP  are presented in Table~\ref{tbl:tdp}. Meaning of the symbols is following:
$M$ stands for the number of vehicles, BKS for the best known solution (to the best of our knowledge, there is no other mTDP solver and BKS is thus the best solution found by one of the evaluated methods), SD is standard deviation, and $T$ is execution time.  
PDB is the percent deviation to BKS of the best solution values found by the algorithm (denoted as $best$), i.e. PDB=($best$-BKS)/BKS.
Similarly, PDM is the percent deviation of the mean solution value to BKS. 
PDBs for the method which found the best solution are highlighted in bold.
Note that {\tt clustering} is a deterministic algorithm, therefore SD is always zero and PDM=PDB and are thus omitted. 

The results show that {\tt proposed} outperforms {\tt k-means} in all cases except six, sometimes by more than $20\%$ in PDB and by more than $30\%$ in PDM. 
On the other hand, {\tt k-means} produces better results for small numbers of robots, where the highest difference of PDM is less than $6\%$.
{\tt clustering} lies between these two. 
Its benefit is much lower computational complexity in comparision to the other methods, with no large gap of produced results to the best found solution.
Difference of computational burden between {\tt k-means} and the other methods increases with an increasing number of vertices as the k-means clustering is much slower that the proposed one.  
This applies especially for higher $M$s, for which the influence of complexity of GRASP is reduced in comparison to {\tt clustering} and {\tt k-means} is thus more than 2 times slower than {\tt proposed}.

The results for mGSP are presented in Table~\ref{tbl:gsp} in the same way as for mTDP.
The running time of all methods slightly increased in comparison to mTDP, but the situation remains the same regarding quality of solutions.
{\tt proposed} produces best results in the majority of cases, followed by {\tt clustering}. 
Again, {\tt k-means} found best solutions for several instances and $M=2$, but the gap does not exceed~$6\%$.

\begin{table*}
\begin{center}
\footnotesize
\begin{tabular}{crrp{0.01em}|rrp{0.01em}|cccrp{0.01em}|cccr}
\toprule
Problem & M & BKS & & \multicolumn{2}{c}{clustering} & & \multicolumn{4}{c}{proposed} & & \multicolumn{4}{c}{k-means}  \\
&~~ & & & PDB \hfil\hfil &  T~[ms]\hfil\hfil & & PDB & PDM & SD\hfil\hfil &  T~[ms]\hfil\hfil &  & PDB & PDM & SD\hfil\hfil &  T~[ms]\hfil\hfil\\
\midrule
\multirow{5}*{berlin52}
 & 2 & 70235 
 & & 3.98 & 0.24
 & & \textbf{0} & 0 & 0.79 & 42.67
 & & 6.82 & 15.76  & 6070.62 & 64.03
\\
 & 4 & 39746 
 & & 0.78 & 0.25
 & & \textbf{0} & 0 & 0& 19.68
 & & 8.84 & 12.77  & 1052.73 & 31.31
\\
 & 6 & 30563 
 & & 0.03 & 0.27
 & & \textbf{0} & 0 & 0& 8.75
 & & 14.24 & 18.07  & 824.95 & 17.48
\\
 & 8 & 25470 
 & & 1.34 & 0.30
 & & \textbf{0} & 0 & 0& 6.61
 & & 13.53 & 21.11  & 1137.95 & 12.69
\\
 & 10 & 23919 
 & & 1.01 & 0.35
 & & \textbf{0} & 0 & 0& 5.14
 & & 14.83 & 19.05  & 742.42 & 8.96
\\
\midrule
\multirow{5}*{bier127}
 & 2 & 2354332 
 & & 10.03 & 0.78
 & & 1.59 & 2.74  & 8388.61 & 396.88
 & & \textbf{0} & 7.66  & 142517.02 & 540.77
\\
 & 4 & 1228367 
 & & 13.10 & 0.64
 & & 5.84 & 5.84  & 0& 116.97
 & & \textbf{0} & 16.83  & 80731.98 & 251.61
\\
 & 6 & 879448 
 & & 4.95 & 0.65
 & & \textbf{0} & 0 & 0& 82.52
 & & 17.52 & 24.22  & 47044.46 & 183.57
\\
 & 8 & 713125 
 & & 0.97 & 0.66
 & & \textbf{0} & 0 & 0& 52.17
 & & 14.96 & 23.25  & 27229.50 & 122.54
\\
 & 10 & 612336 
 & & 0.35 & 0.73
 & & \textbf{0} & 0 & 0& 38.65
 & & 19.48 & 26.01  & 18738.51 & 99.08
\\
\midrule
\multirow{5}*{gil262}
 & 2 & 153716 
 & & 8.66 & 3.23
 & & \textbf{0} & 0.98  & 601.31 & 3055.23
 & & 0.06 & 4.32  & 3130.36 & 4342.95
\\
 & 4 & 87114 
 & & 6.96 & 2.16
 & & \textbf{0} & 0.21  & 81.67 & 1019.36
 & & 12.65 & 13.30  & 209.34 & 2187.43
\\
 & 6 & 65428 
 & & 3.10 & 2.07
 & & \textbf{0} & 0.04  & 22.76 & 599.17
 & & 13.56 & 18.98  & 2124.00 & 1168.37
\\
 & 8 & 55306 
 & & 1.90 & 2.10
 & & \textbf{0} & 0.01  & 8.14 & 447.16
 & & 16.25 & 21.84  & 1566.33 & 776.67
\\
 & 10 & 50292 
 & & 0.75 & 2.18
 & & \textbf{0} & 0 & 2.63 & 329.39
 & & 19.43 & 23.64  & 1417.02 & 617.23
\\
\midrule
\multirow{5}*{lin318}
 & 2 & 3140312 
 & & 15.07 & 4.29
 & & 4.55 & 5.75  & 15004.90 & 5855.93
 & & \textbf{0} & 2.92  & 32535.88 & 5931.95
\\
 & 4 & 1811206 
 & & 6.46 & 2.92
 & & \textbf{0} & 0.24  & 2817.56 & 1644.35
 & & 1.80 & 7.44  & 36532.68 & 2239.29
\\
 & 6 & 1431946 
 & & 4.18 & 2.86
 & & \textbf{0} & 0.08  & 1004.02 & 940.59
 & & 2.35 & 5.92  & 31411.96 & 1371.37
\\
 & 8 & 1214427 
 & & 3.24 & 2.85
 & & \textbf{0} & 0.03  & 292.51 & 563.80
 & & 4.56 & 8.13  & 44304.42 & 976.14
\\
 & 10 & 1129348 
 & & 3.93 & 3.03
 & & \textbf{0} & 0.01  & 256.28 & 565.05
 & & 3.55 & 6.49  & 27180.85 & 760.35
\\
\midrule
\multirow{5}*{pcb442}
 & 2 & 5292831 
 & & 5.07 & 9.13
 & & \textbf{0} & 0.56  & 17515.61 & 14006.46
 & & 5.56 & 6.46  & 26968.07 & 24343.18
\\
 & 4 & 2849704 
 & & 4.96 & 6.08
 & & \textbf{0} & 0.44  & 7294.97 & 3758.14
 & & 14.27 & 14.61  & 6929.25 & 8974.81
\\
 & 6 & 2055340 
 & & 2.07 & 5.75
 & & \textbf{0} & 0.14  & 2272.75 & 1641.71
 & & 17.98 & 21.26  & 33959.70 & 4861.82
\\
 & 8 & 1607307 
 & & 1.65 & 5.73
 & & \textbf{0} & 0.11  & 1234.00 & 1251.63
 & & 26.51 & 30.42  & 30783.56 & 3115.88
\\
 & 10 & 1402893 
 & & 2.74 & 5.84
 & & \textbf{0} & 0.05  & 898.36 & 988.54
 & & 29.37 & 33.29  & 24376.01 & 2261.04
\\
\midrule
\multirow{5}*{rat575}
 & 2 & 994166 
 & & 7.01 & 16.96
 & & \textbf{0} & 0.96  & 3631.90 & 37961.41
 & & 2.20 & 3.10  & 3053.47 & 66531.58
\\
 & 4 & 524957 
 & & 6.65 & 10.47
 & & \textbf{0} & 0.50  & 1110.60 & 9644.29
 & & 9.78 & 13.40  & 12210.99 & 23472.77
\\
 & 6 & 375842 
 & & 6.60 & 10.20
 & & \textbf{0} & 0.22  & 375.69 & 4100.55
 & & 11.89 & 13.89  & 8107.88 & 11401.33
\\
 & 8 & 303082 
 & & 3.46 & 10.10
 & & \textbf{0} & 0.06  & 137.15 & 2312.13
 & & 15.63 & 17.79  & 5449.48 & 7426.09
\\
 & 10 & 260937 
 & & 1.37 & 10.47
 & & \textbf{0} & 0.05  & 101.79 & 1406.92
 & & 18.82 & 22.19  & 4480.78 & 5577.58
\\
\midrule
\multirow{5}*{u724}
 & 2 & 7479059 
 & & 8.28 & 26.90
 & & \textbf{0} & 1.31  & 39681.60 & 57579.28
 & & 2.86 & 3.47  & 24549.00 & 79037.02
\\
 & 4 & 3904506 
 & & 7.77 & 17.47
 & & \textbf{0} & 0.52  & 8792.00 & 16542.24
 & & 9.23 & 12.04  & 59939.66 & 26556.46
\\
 & 6 & 2926359 
 & & 5.31 & 16.85
 & & \textbf{0} & 0.38  & 4477.46 & 8135.85
 & & 10.39 & 11.53  & 22159.80 & 15149.82
\\
 & 8 & 2427262 
 & & 4.49 & 17.03
 & & \textbf{0} & 0.11  & 1496.06 & 6363.60
 & & 11.95 & 13.89  & 28998.38 & 10828.30
\\
 & 10 & 2111492 
 & & 2.46 & 17.42
 & & \textbf{0} & 0.12  & 1014.74 & 3725.07
 & & 14.90 & 16.67  & 34886.45 & 8564.65
\\
\midrule
\multirow{5}*{pr1002}
 & 2 & 65541078 
 & & 5.99 & 52.28
 & & 0.35 & 0.99  & 215703.18 & 183435.59
 & & \textbf{0} & 0.63  & 227427.16 & 279021.00
\\
 & 4 & 36553749 
 & & 13.90 & 35.78
 & & 1.69 & 2.20  & 85095.75 & 53400.13
 & & \textbf{0} & 0.75  & 192716.60 & 81455.54
\\
 & 6 & 26676866 
 & & 9.48 & 36.38
 & & \textbf{0} & 0.28  & 33816.36 & 25943.29
 & & 1.17 & 2.10  & 318901.94 & 43332.34
\\
 & 8 & 20946133 
 & & 7.57 & 36.03
 & & \textbf{0} & 0.18  & 20700.88 & 14389.91
 & & 8.59 & 10.53  & 544729.91 & 30745.18
\\
 & 10 & 19540052 
 & & 6.21 & 37.82
 & & 0.96 & 1.14  & 17698.56 & 9307.68
 & & \textbf{0} & 3.34  & 524410.71 & 22968.16
\\
\midrule
\end{tabular}
\end{center}
\vspace{-0.2em}
\caption{Comparison of the algorithms for mTDP.\vspace{-0.4em}}
\label{tbl:tdp}
\end{table*}

\begin{table*}
\begin{center}
\footnotesize
\begin{tabular}{crrp{0.01em}|rrp{0.01em}|cccrp{0.01em}|cccr}
\toprule
Problem & M & BKS & & \multicolumn{2}{c}{clustering} & & \multicolumn{4}{c}{proposed} & & \multicolumn{4}{c}{k-means}  \\
&~~ & & & PDB \hfil\hfil &  T~[ms]\hfil\hfil & & PDB & PDM & SD\hfil\hfil &  T~[ms]\hfil\hfil &  & PDB & PDM & SD\hfil\hfil &  T~[ms]\hfil\hfil\\
\midrule
\multirow{5}*{berlin52}
 & 2 & 1289.3665 
 & & 15.49 & 0.29
 & & 2.55 & 2.55  & 0.06 & 55.73
 & & \textbf{0} & 7.67  & 101.55 & 65.39
\\
 & 4 & 711.9565 
 & & 8.92 & 0.32
 & & \textbf{0} & 0 & 0& 24.22
 & & 5.10 & 9.70  & 23.96 & 40.98
\\
 & 6 & 569.3354 
 & & 1.28 & 0.33
 & & \textbf{0} & 0 & 0& 11.74
 & & 8.06 & 11.66  & 12.52 & 21.95
\\
 & 8 & 510.9161 
 & & 2.82 & 0.36
 & & \textbf{0} & 0 & 0& 8.35
 & & 4.91 & 9.25  & 14.94 & 14.45
\\
 & 10 & 462.7143 
 & & 0& 0.38
 & & \textbf{0} & 0 & 0& 4.53
 & & 9.43 & 13.30  & 16.38 & 10.43
\\
\midrule
\multirow{5}*{bier127}
 & 2 & 16938.3426 
 & & 16.94 & 0.92
 & & \textbf{0} & 0.58  & 59.14 & 465.85
 & & 1.19 & 9.68  & 1212.99 & 613.18
\\
 & 4 & 9015.8078 
 & & 10.15 & 0.78
 & & \textbf{0} & 0.11  & 7.17 & 179.04
 & & 6.09 & 21.01  & 774.75 & 301.05
\\
 & 6 & 6439.1727 
 & & 5.06 & 0.81
 & & \textbf{0} & 0 & 0.25 & 112.68
 & & 14.08 & 23.26  & 368.25 & 221.56
\\
 & 8 & 5297.5682 
 & & 2.33 & 0.86
 & & \textbf{0} & 0 & 0& 63.20
 & & 10.91 & 22.10  & 280.29 & 153.78
\\
 & 10 & 4517.5042 
 & & 1.83 & 0.94
 & & \textbf{0} & 0 & 0& 42.99
 & & 15.53 & 24.25  & 161.22 & 120.06
\\
\midrule
\multirow{5}*{gil262}
 & 2 & 557.1953 
 & & 12.84 & 3.46
 & & \textbf{0} & 1.03  & 2.38 & 3465.02
 & & 0.39 & 3.18  & 11.40 & 4415.25
\\
 & 4 & 324.9695 
 & & 8.78 & 2.51
 & & \textbf{0} & 0.30  & 0.49 & 1028.91
 & & 9.60 & 10.06  & 1.77 & 2628.16
\\
 & 6 & 241.1403 
 & & 5.21 & 2.49
 & & \textbf{0} & 0.12  & 0.16 & 604.07
 & & 13.62 & 17.69  & 6.33 & 1410.60
\\
 & 8 & 208.0251 
 & & 4.62 & 2.57
 & & \textbf{0} & 0.03  & 0.06 & 378.99
 & & 15.12 & 20.39  & 6.27 & 1009.05
\\
 & 10 & 195.8603 
 & & 1.75 & 2.89
 & & \textbf{0} & 0 & 0& 315.60
 & & 13.94 & 18.16  & 4.62 & 788.35
\\
\midrule
\multirow{5}*{lin318}
 & 2 & 9729.8214 
 & & 17.30 & 4.85
 & & 5.48 & 6.23  & 40.37 & 6412.44
 & & \textbf{0} & 1.41  & 46.65 & 6757.29
\\
 & 4 & 5614.8969 
 & & 9.26 & 3.67
 & & \textbf{0} & 0.40  & 8.24 & 1769.25
 & & 1.00 & 5.74  & 112.89 & 2510.83
\\
 & 6 & 4390.3071 
 & & 4.68 & 3.52
 & & \textbf{0} & 0.16  & 3.95 & 878.58
 & & 2.62 & 7.51  & 118.88 & 1773.18
\\
 & 8 & 3728.5347 
 & & 4.50 & 3.77
 & & \textbf{0} & 0.02  & 0.75 & 606.11
 & & 5.42 & 9.02  & 135.38 & 1315.97
\\
 & 10 & 3504.8153 
 & & 4.69 & 4.13
 & & \textbf{0} & 0.01  & 0.43 & 547.21
 & & 2.99 & 5.94  & 96.50 & 1001.21
\\
\midrule
\multirow{5}*{pcb442}
 & 2 & 11670.8481 
 & & 11.20 & 10.06
 & & 3.23 & 4.25  & 45.68 & 13903.57
 & & \textbf{0} & 3.06  & 155.85 & 19525.87
\\
 & 4 & 6076.4795 
 & & 5.80 & 7.07
 & & \textbf{0} & 0.39  & 10.10 & 3348.96
 & & 14.86 & 15.81  & 22.50 & 8445.16
\\
 & 6 & 4438.6460 
 & & 4.89 & 6.70
 & & \textbf{0} & 0.19  & 3.57 & 1725.01
 & & 17.89 & 21.13  & 70.81 & 5055.13
\\
 & 8 & 3634.2044 
 & & 3.33 & 7.01
 & & \textbf{0} & 0.05  & 1.21 & 1113.55
 & & 22.60 & 24.91  & 61.11 & 3643.70
\\
 & 10 & 3197.8358 
 & & 3.34 & 7.64
 & & \textbf{0} & 0.01  & 0.39 & 733.25
 & & 24.93 & 28.05  & 47.30 & 2869.68
\\
\midrule
\multirow{5}*{rat575}
 & 2 & 1624.6539 
 & & 12.92 & 17.25
 & & \textbf{0} & 1.14  & 8.45 & 29891.17
 & & 0.50 & 1.17  & 5.19 & 50161.28
\\
 & 4 & 885.9464 
 & & 12.11 & 11.67
 & & \textbf{0} & 0.46  & 2.41 & 8358.31
 & & 4.77 & 8.81  & 20.46 & 20297.80
\\
 & 6 & 647.9221 
 & & 5.97 & 11.25
 & & \textbf{0} & 0.33  & 1.11 & 4596.77
 & & 5.93 & 8.06  & 14.37 & 11989.79
\\
 & 8 & 527.9039 
 & & 5.70 & 11.55
 & & \textbf{0} & 0.14  & 0.32 & 2678.61
 & & 9.93 & 11.43  & 7.55 & 8496.11
\\
 & 10 & 444.0691 
 & & 3.56 & 12.76
 & & \textbf{0} & 0.10  & 0.21 & 1842.64
 & & 15.72 & 19.20  & 7.37 & 6800.49
\\
\midrule
\multirow{5}*{u724}
 & 2 & 10053.0689 
 & & 20.39 & 27.42
 & & 1.63 & 3.08  & 47.51 & 64240.43
 & & \textbf{0} & 0.85  & 31.56 & 87189.09
\\
 & 4 & 5492.4174 
 & & 12.34 & 18.58
 & & \textbf{0} & 0.52  & 14.49 & 16706.83
 & & 2.46 & 5.66  & 95.87 & 31098.84
\\
 & 6 & 3967.1502 
 & & 9.06 & 17.38
 & & \textbf{0} & 0.35  & 4.86 & 10488.02
 & & 7.99 & 9.06  & 39.43 & 18807.40
\\
 & 8 & 3377.1366 
 & & 6.68 & 18.08
 & & \textbf{0} & 0.14  & 2.61 & 6682.94
 & & 8.28 & 9.73  & 24.42 & 13853.41
\\
 & 10 & 2805.0308 
 & & 4.43 & 20.14
 & & \textbf{0} & 0.10  & 1.27 & 4474.89
 & & 16.32 & 17.96  & 30.07 & 10550.73
\\
\midrule
\multirow{5}*{pr1002}
 & 2 & 60325.4178 
 & & 17.21 & 54.04
 & & \textbf{0} & 1.22  & 266.22 & 242200.87
 & & 1.78 & 3.06  & 289.47 & 271763.71
\\
 & 4 & 34423.7763 
 & & 20.83 & 36.08
 & & 5.10 & 5.62  & 98.06 & 52536.92
 & & \textbf{0} & 0.99  & 137.73 & 77147.53
\\
 & 6 & 25056.6643 
 & & 12.34 & 34.25
 & & \textbf{0} & 0.32  & 39.35 & 33415.13
 & & 3.26 & 4.06  & 236.43 & 49881.61
\\
 & 8 & 20813.4050 
 & & 9.51 & 34.61
 & & \textbf{0} & 0.24  & 23.37 & 19364.56
 & & 4.46 & 6.70  & 456.05 & 35652.01
\\
 & 10 & 17750.8518 
 & & 9.21 & 38.10
 & & \textbf{0} & 0.20  & 15.70 & 11952.95
 & & 5.79 & 8.58  & 491.02 & 25752.29
\\
\midrule
\end{tabular}
\end{center}
\vspace{-0.2em}
\caption{Comparison of the algorithms for mGSP.\vspace{-0.4em}}
\label{tbl:gsp}
\end{table*}

\section{Conclusion}
\label{sec:conclusion}
We formulated mutli-vehicle variants of two routing problems -- the Traveling Deliveryman Problem and the Graph Search Problem and introduced a novel clustering approach which is designed especially for these two problems.
The proposed clustering was then used together with the GRASP metaheuristics in the cluster-first route-second scheme as an integrated approach to the problems.
The proposed integrated approach is, based on the performed experimental results, suitable to solve both problems as it outperforms the standard approach based on k-means clustering in quality of found solutions in the majority of cases with lower computational burden. 
Moreover, the proposed clustering can be used as a standalone solver as it produces solutions slightly worse than the integrated approach, but substantially faster.

Future research will go in two directions.
We would like to use the proposed clustering in GRASP as a constructive heuristic and together with design of neighborhoods for exchanging vertices between routes extend pure GRASP for mTDP and mGSP.   
The second stream will focus on application of the proposed approach in other robotic applications. 
Similarly to using a mTSP solver for the exploration task~\cite{Faigl2012} and a GSP solver for search in a priori unknown environment~\cite{Kulich2014, Kulich2016}, we would like to study properties of mGSP for multi-robot search in an unknown space.

\section*{Acknowledgments}
This work has been supported by the European Union's Horizon 2020 research and innovation programme under grant agreement No 688117 and the Technology Agency of the Czech Republic under the project no.~TE01020197 \enquote{Centre for Applied Cybernetics}.

\bibliographystyle{IEEEtran}
\bibliography{main}

\end{document}